%% file: root.tex
\documentclass[letterpaper, 10 pt, conference]{ieeeconf}  

\IEEEoverridecommandlockouts                              

\usepackage{graphicx} 
\usepackage{amsmath} 
\usepackage{amssymb}  
\usepackage{amsbsy}
\newcommand{\say}[1]{\emph{#1}}

\usepackage[textsize=footnotesize]{todonotes}
\usepackage{verbatim}

\usepackage{algorithm}
\usepackage{algpseudocode}

\usepackage[utf8]{inputenc}

\usepackage{cite}
\usepackage{subcaption}
\usepackage{booktabs}
\usepackage{tikz}
\usepackage{pgfplots}

\let\labelindent\relax
\usepackage{enumitem}

\usepackage[nolist]{acronym}

\pgfplotsset{grid style={dotted, gray}}
\pgfplotsset{minor grid style={dotted,gray}}
\pgfplotsset{every tick label/.append style={font=\tiny}}
\pgfplotsset{every axis/.append style={font=\small}}
\pgfplotsset{ylabel near ticks}
\pgfplotsset{xlabel near ticks}
\newlength\figureheight 
\newlength\figurewidth

\input{acronyms.tex}

\title{\LARGE \bf
Improving dual-arm assembly by master-slave compliance
}

\author{Markku Suomalainen\textsuperscript{1,2}, Sylvain Calinon\textsuperscript{3}, Emmanuel Pignat\textsuperscript{3} and Ville Kyrki\textsuperscript{2}
\thanks{This work was supported by Academy of Finland, decision 286580.}
\thanks{\textsuperscript{1} M.\ Suomalainen is currently with University of Oulu, Finland {\tt\small markku.suomalainen@oulu.fi}}
\thanks{\textsuperscript{2}School of Electrical Engineering, Aalto University, Finland {\tt\small ville.kyrki@aalto.fi}}
\thanks{\textsuperscript{3} Idiap Research Institute, Switzerland {\tt\small firstname.surname@idiap.ch}}}

\begin{document}

\maketitle
\thispagestyle{empty}
\pagestyle{empty}

\begin{abstract}
In this paper we show how different choices regarding compliance affect a dual-arm assembly task. In addition, we present how the compliance parameters can be learned from a human demonstration. Compliant motions can be used in assembly tasks to mitigate pose errors originating from, for example, inaccurate grasping. We present analytical background and accompanying experimental results on how to choose the center of compliance to enhance the convergence region of an alignment task. Then we present the possible ways of choosing the compliant axes for accomplishing alignment in a scenario where orientation error is present. We show that an earlier presented Learning from Demonstration method can be used to learn motion and compliance parameters of an impedance controller for both manipulators. The learning requires a human demonstration with a single teleoperated manipulator only, easing the execution of demonstration and enabling usage of manipulators at difficult locations as well. Finally, we experimentally verify our claim that having both manipulators compliant in both rotation and translation can accomplish the alignment task with less total joint motions and in shorter time than moving one manipulator only. In addition, we show that the learning method produces the parameters that achieve the best results in our experiments.
\end{abstract}

\section{INTRODUCTION}
\label{sec:intro}
\begin{figure}[tb]
	\centering
	\includegraphics[width=.9\columnwidth]{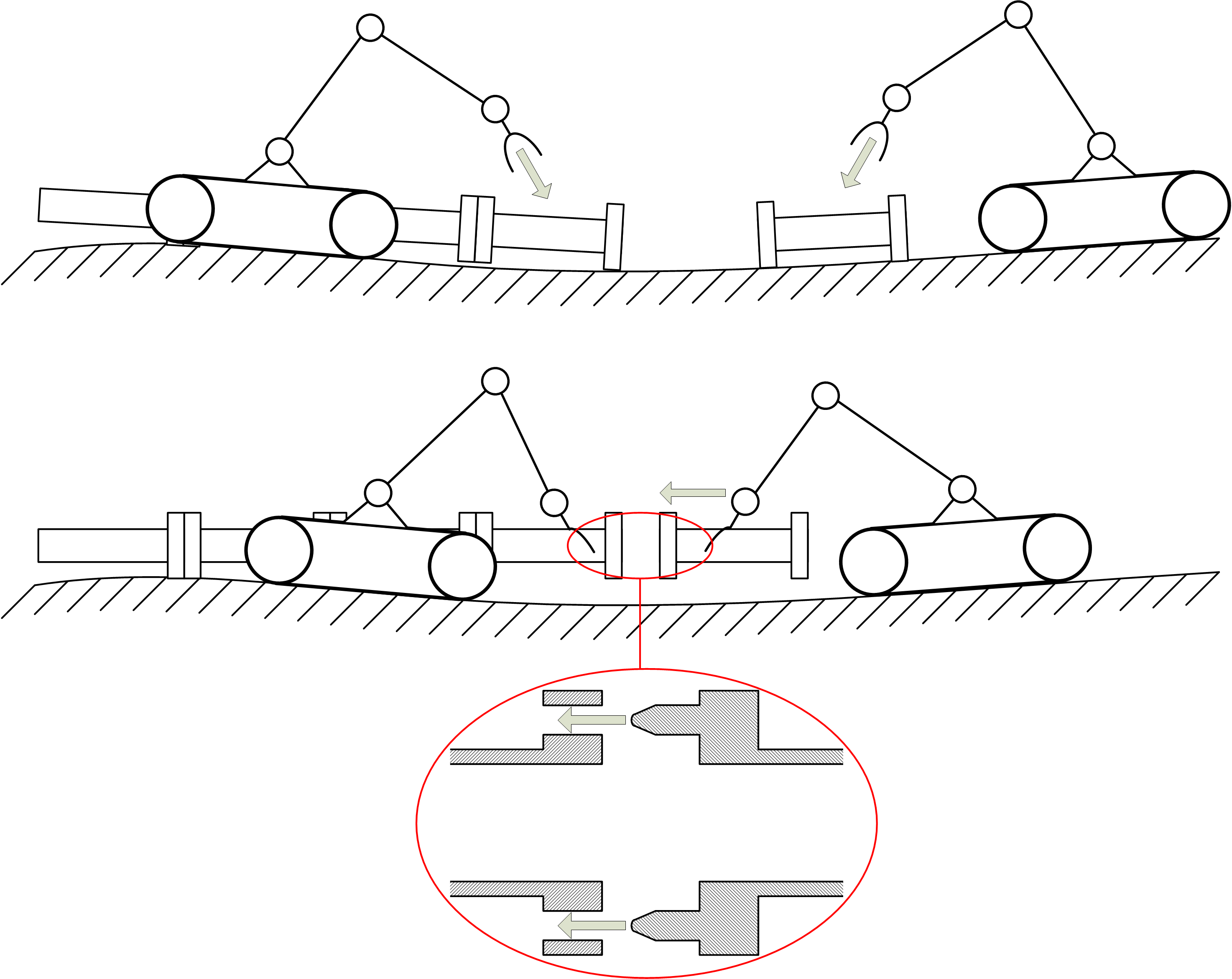}
	\caption{\small{An illustration of a heavy-duty assembly task with two manipulators grasping workpieces and taking advantage of compliance (source: \cite{suomalainen2016})} }
	\label{fig:assembly}
\vspace{-1em}
\end{figure}

When a human performs alignment motions as depicted in Fig.~\ref{fig:assembly}, often only one hand moves the piece actively and the other hand only complies with the motions to ease the alignment. Whenever there is not a fixture available to hold one of the workpieces, they must both be held to perform alignment. Therefore, to allow low-threshold automation of currently human-performed tasks, robots must be able to execute these tasks similarly to humans and learn the tasks efficiently~\cite{smith2012dual}.

When a dual-arm robot needs to grasp and assemble two workpieces, their relative position and orientation will differ in each consecutive trials. This issue is independent of the controller being centralized or decentralized because location of a grasp on a workpiece is always limited in precision. Pose errors during alignment caused by this uncertainty are difficult to mitigate with a vision system due to contact between the workpieces and occlusions. Another approach to mitigate these errors is leveraging compliance, i.e.~take advantage of the workpieces geometry to slide them into contact and, furthermore, into alignment. For this kind of tasks, impedance control \cite{hogan1985impedance} is a useful tool due to its ability to perform both free-space and in-contact motions. Even though compliant motions for dual-arm manipulation can be achieved through pre-planned trajectories \cite{caccavale2001achieving}, this can be a strenuous task.

For efficiently teaching robots new tasks, \ac{lfd} is a well established paradigm \cite{argall2009survey} where the idea is that the user can simply show the robot the required task, which is often more efficient than designing trajectories. Teaching a dual-arm manipulator using \ac{lfd}, however, poses certain challenges. Especially when using kinesthetic teaching, often perceived as the easiest methods by users \cite{fischer2016comparison}, it can be difficult to simultaneously guide two manipulators due to friction at the joints, kinematic redundancy and inertia of the manipulator. Furthermore, kinesthetic teaching cannot be used with heavy robots (Fig.~\ref{fig:assembly}) or in dangerous environments, restricting the use of methods limited to kinesthetic teaching only.


We propose to use linear translational motions in one arm along with compliance in both arms to perform dual-arm alignment tasks under pose uncertainty. We first present an analysis of how rotational and translational compliances affect the alignment task. We also show how important the choice of \ac{coc} is for a successful alignment. Then we present how this kind of motions for an impedance controller can be efficiently learned from human demonstrations with an \ac{lfd} method presented in \cite{suomalainen2017, suomalainen2018learn6D}. Finally, we explain how compliances for both arms can be learned from a teleoperated demonstration where one arm is fixed, thus simplifying the demonstration. With the learned compliances we can effectively use contact forces to guide the workpieces during alignment.

In this work we present the following contributions: 1) we present an analysis on the importance of choosing the \ac{coc} and compliant axes in an assembly task; 2) we show that the compliance parameters learned by the method in \cite{suomalainen2018learn6D} can be used in both arms performing a dual-arm assembly task, even from demonstration with one arm only; and 3) we show through experiments that having compliance on both manipulators eases an assembly task by decreasing the time and joint motions required to complete the task.



\section{RELATED WORK}
\label{sec:related}
There exist several works on leveraging geometry in single- and dual-arm assembly. 
Verscheure et al.~\cite{verscheure2008line} showed how probing motions can be used to deduce geometry of a task and Karayiannidis et al.~\cite{karayiannidis2014online} presented how to calibrate tools when facing pose uncertainties due to grasping. Almeida and Karayiannidis~\cite{almeida2016folding} performed dual-arm folding assembly by explicitly computing the contact point kinematics and developed a controller based on feedback linearisation. Whereas these papers consider similar problems as we do, the difference is the simplicity: in this paper we show how to maximize the power of simple linear motions with compliance and also how a lay user could teach them to a robot efficiently.


Learning from Demonstration for dual-arm robots has been a topic for research for more than ten years \cite{zollner2004programming}. Silv{\'e}rio et al.~\cite{silverio2015learning} showed how to learn the full pose for a task-parametrized dual-arm task. Recently, Batinica et al.~\cite{batinica2017compliant} demonstrated how Compliant Movement Primitives \cite{denivsa2016learning} can be used to learn from human demonstration dual-arm motions requiring compliance by combining desired joint motion trajectories and corresponding joint torque signals. Whereas this method can learn more complex trajectories in free space, the tight temporal coupling between pose and torque trajectories makes it susceptible to pose errors. As the scope of this paper is in-contact alignment tasks, we prefer error tolerance over complexity of applicable motions and thus base our work on the method from \cite{suomalainen2018learn6D}.

Recently there has been an increasing interest in learning from teleoperated demonstrations. As shown by Fischer et al.~\cite{fischer2016comparison}, this requires more error tolerance from the learning method. Pervez et al.~\cite{pervez2017novel} overcame this problem by manually choosing one good demonstration as a baseline. Havoutis and Calinon \cite{havoutis18AURO} looked into mixed teleoperation, where the human would first guide the robot through teleoperation, but the robot could then finish the task autonomously. Thus, a method that can learn from a teleoperated demonstration will have a broader range of applications than learning from kinesthetic teaching only.

A strategy for dual-arm peg-in-hole assembly with a compliant manipulator was presented by Zhang et al.~\cite{zhang2017peg}. They proposed a hand-crafted two-phase method to overcome positional inaccuracies, creating a sequential search using feedback from a \ac{ft} sensor. Their observation from a human performing the task was that after initiating contact (i.e.~their second phase, which corresponds to the work in this paper), the human mainly adjusts the orientation of the workpieces. This enhances our intuition that for a dual-arm peg-in-hole task, actively moving both arms does not necessarily improve completion of the task compared to having a single-arm operation. However, we observe that a human performing the peg-in-hole task with two arms allows the passive arm to move in both orientation and translation according to the motions of the active arm moving the peg. This motivates our research on the properties of dual-arm alignment task with rotational and translational compliance on both arms.

\section{METHOD}
\label{sec:method}
We consider a scenario where a \say{master} manipulator is moved actively along a linear trajectory in the Cartesian space. The master can also be compliant (defined as a stiffness of 0 in this paper) in rotation and/or translation, in which case environmental forces can displace it from the linear trajectory. The \say{slave} manipulator is commanded to stay stationary, but may be set compliant along chosen axes and thus move because of contact forces. We assume both manipulators are grasping their respective workpieces to be aligned and that the uncertainties regarding the location of the workpieces tooltips are low enough such that the workpieces can be brought into contact. In addition, we assume the \ac{coc} of each manipulator match the \ac{tcs} and can be set arbitrarily. We do not require the manipulators to be centrally controlled.

An illustration of a peg-in-hole scenario meeting the preceding requirements is shown in Fig.~\ref{fig:failmodes}. In this scenario the task is to insert the peg into the hole by moving the right master manipulator along the direction of the tool, the z-axis in \ac{tcs} as shown in Fig.~\ref{fig:failmodes}. 

Methods presented in this paper intend to make the convergence region of the manipulators, i.e.~the margin of error from which the alignment is accomplished, as large as possible. In addition, we investigate whether there are additional benefits for using dual-arm compliant robots for assembly, such as smaller workspace requirements. In Section~\ref{sec:pih} we observe how different compliance parameters can be set to enlarge the convergence region and decrease the required joint motions. In Section~\ref{sec:LfD} we shortly present the method from \cite{suomalainen2018learn6D} for learning compliant behaviour from human demonstration and how it is applied to a dual-arm scenario. Finally, we also show how to control the manipulators during reproduction of the motion.

\subsection{Mechanics of compliant dual-arm assembly}
\label{sec:pih}

\begin{figure}
        \centering
        \begin{subfigure}[b]{0.325\textwidth}
            \centering
            \includegraphics[width=\textwidth]{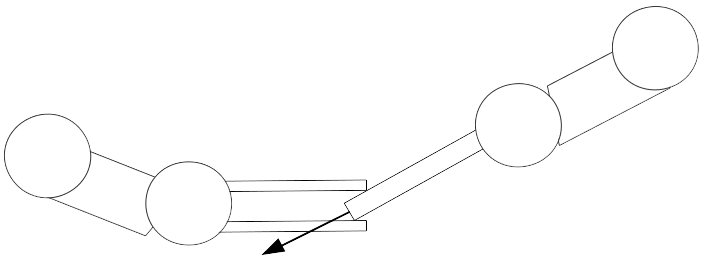}
            \caption{Starting pose and possible jamming}
        \end{subfigure}
        \begin{subfigure}[b]{0.225\textwidth}
            \centering
            \includegraphics[width=\textwidth]{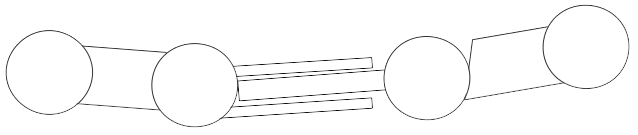}
            \caption{Successful alignment}
        \end{subfigure}
        \hfill
        \begin{subfigure}[b]{0.225\textwidth}  
            \centering 
            \includegraphics[width=\textwidth]{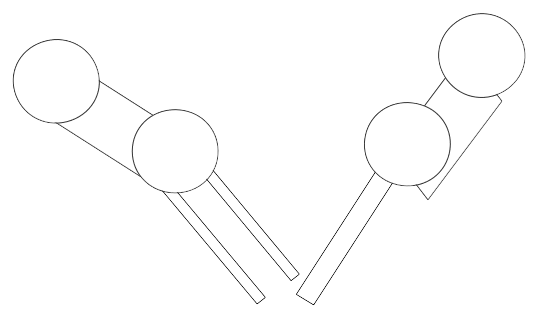}
            \caption{Rotation failure}
        \end{subfigure}
        \caption{\small{Possible outcomes of dual-arm peg-in-hole with compliance. The right manipulator is the master, moving along the direction of the arrow, and the left one is the slave.}}
        \label{fig:failmodes}
\end{figure}

When performing dual-arm assembly with compliance, there are a number of possible failure types and variables that affect them. 
In \cite{whitney1982quasi} there is a thorough analysis on two possible insertion failures, jamming and wedging, which depend on the geometry and forces affecting the peg. We assume that our workpieces cannot be deformed, which is required in wedging, but jamming can occur with high friction and high error in orientation, as explained in \cite{whitney1982quasi}. However, when also the hole is compliant, the problem becomes different. If there is enough friction the peg can still be jammed, but there is also a new failure mode, illustrated in Fig.~\ref{fig:failmodes}c.

Considering the figure, when the peg is moved along the direction of the arrow and both master and slave are compliant, there are three possible outcomes: (a) jamming, which means that the peg does not move at all or very little, (b) the alignment is completed and (c) both workpieces start rotating in the wrong direction and the task fails. Besides the geometry of the workpieces, this behaviour depends on three factors: 1) center of compliance, 2) axes of compliance and 3) the orientation error between the peg and the hole. As factor 3) depends on the setup, it is considered only in a comparative manner in Section~\ref{sec:experiments}. In the next sections we consider the effect of the first two factors. 

\subsubsection{Center of Compliance}
\label{sec:coc}
\begin{figure}
        \centering
        \begin{subfigure}[b]{0.325\textwidth}
            \centering
            \includegraphics[width=\textwidth]{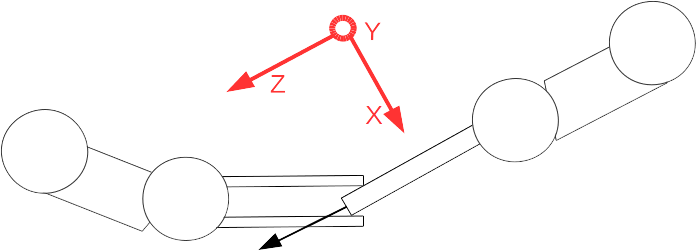}
            \caption{Starting pose and direction of motion}
        \end{subfigure}
        \begin{subfigure}[b]{0.225\textwidth}
            \centering
            \includegraphics[width=\textwidth]{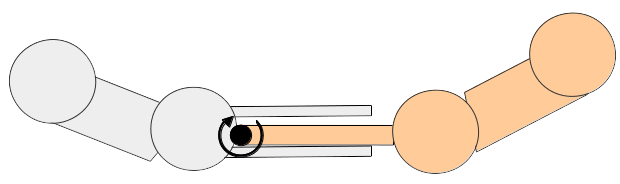}
            \caption{Master compliant at tooltip}
        \end{subfigure}
        \hfill
        \begin{subfigure}[b]{0.225\textwidth}  
            \centering 
            \includegraphics[width=\textwidth]{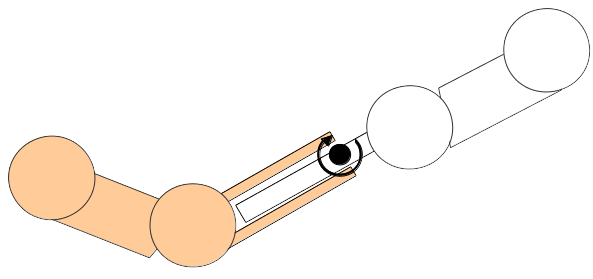}
            \caption{Slave compliant at tooltip}
        \end{subfigure}
        \vskip\baselineskip
        \begin{subfigure}[b]{0.225\textwidth}   
            \centering 
            \includegraphics[width=\textwidth]{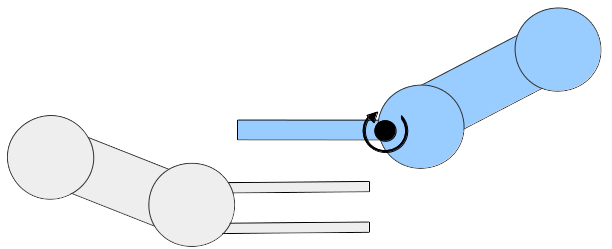}
            \caption{Master compliant at wrist}
        \end{subfigure}
        \quad
        \begin{subfigure}[b]{0.225\textwidth}   
            \centering 
            \includegraphics[width=\textwidth]{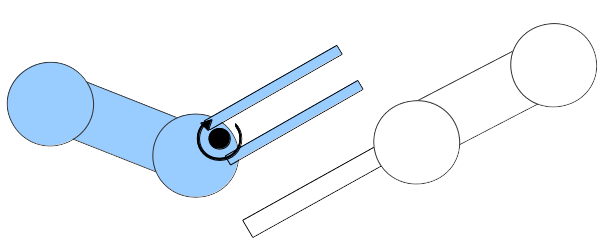}
            \caption{Slave compliant at wrist}
        \end{subfigure}
        \caption{\small Illustration of the relevance of choosing the \ac{coc} ($\bullet$). The orange manipulators have the \ac{coc} at tooltip and the blue manipulators at the wrist. (a) shows the starting poses of manipulators of each case, along with the direction the master is commanded to move and the coordinate frame used throughout this paper. Figures on the left, (b) and (d), represent the end result of the master having rotational compliance at different \ac{coc}s. Figures on the right((c),(e)) represent the end result of the slave having rotational compliance at different \ac{coc}s. The blue manipulators would not actually move due to physical constraints, but the figures show how they could rotate without the constraints.} 
        \label{fig:coc}
    \end{figure}

\ac{coc} is the point where the stiffness matrix is diagonal in coordinates whose origin is at the \ac{coc} \cite{whitney1982quasi}. This means that both the control commands and the axes of compliance are defined in a coordinate system at this point. Using Fig.~\ref{fig:coc}, we investigate two possible choices: having the \ac{coc} at the wrist of the manipulator (blue manipulators) or having it at the tooltip (orange manipulators).

We first analyse the scenario where the slave manipulator is fixed (gray colour), i.e.~the setting is similar to a single-arm peg-in-hole with the master manipulator moving the peg forward as explained earlier, while the master manipulator is rotationally compliant in either tooltip (Fig.~\ref{fig:coc}b) or wrist (Fig.~\ref{fig:coc}d) but there is no translational compliance. The key point is to observe that in the starting scenario (Fig.~\ref{fig:coc}a), the origin of the TCS at tooltip is already in the hole and thus the rotational compliance is enough to align the master tool with the slave, whereas the TCS at the wrist would require compliance in translation along x-axis to align the tools. However, as there is no translational compliance in this scenario, the tool of the master can only rotate to the position shown in Fig.~\ref{fig:coc}d. As this is not physically possible, the motion would result in jamming and a failed task. 
 
The results are identical if the master is stiff (i.e.~motion but no compliance) and the slave manipulator is set to rotationally compliant. Fig.~\ref{fig:coc}c shows the slave manipulator rotating around the \ac{coc} at tooltip caused by the external forces of the master such that alignment is possible. Rotation around the wrist of the slave manipulator would result in the pose of Fig.~\ref{fig:coc}e, and thus jamming and a failed task. This analysis is further examined in the experiments in Section \ref{sec:experiments}. We note that with translational compliance the alignment would succeed with either location  of the \ac{coc}, but the fact that rotational compliance at tooltip is more efficient is expected to have an effect on the results.

\subsubsection{Axes of compliance}

Starting again from the scenario of Fig.~\ref{fig:coc}a, without any compliance it is impossible to perform alignment. As discussed in Section~\ref{sec:coc}, with a correctly chosen \ac{coc}, rotational compliance only is sufficient for completing the alignment. However, the results in \cite{suomalainen2018learn6D} show that adding translational compliance increases the convergence region for a single-arm task. As our goal is to maximize the convergence region and consider methods that accomplish the alignment regardless of the choice of \ac{coc}, the option of single manipulator rotation-only compliant will not be considered further.

We consider the most prominent types of compliance for achieving the goal regardless of the choice of \ac{coc} to be:

\begin{itemize}
\item compliance in translation and rotation in the master only (e.g.~single-arm peg-in-hole, the exact setting from peg-in-hole experiments in~\cite{suomalainen2018learn6D}),
\item compliance in translation and rotation in the slave only,
\item compliance in translation and rotation in both robots,
\item compliance in rotation only in both robots.
\end{itemize}

There exist several other combinations in which the task can succeed as well, such as both translational and rotational compliance in one manipulator and either translational or rotational compliance on the other. However, we will focus on the four itemized cases as they can be considered the corner cases, and thus we expect them to present the most interesting results.

The first case is essentially a single-arm assembly and serves as a baseline. In the second case we explore if there is difference whether the compliance is in the master or in the slave. In the third option, we expect to see an improvement over first and second option, since now both robots can contribute to the alignment. The fourth option is a theoretically interesting one. As mentioned, one robot being rotationally compliant only at the wrist is insufficient to complete the defined task, but with both robots being compliant in rotation the missing compliance in translation is compensated even if \ac{coc} is chosen at the wrist. In contrast, having only translational compliance in both robots is not enough to complete the motion in any \ac{coc} unless the rotations are already aligned. These cases are experimentally studied in Section~\ref{sec:experiments}.

\subsection{Control and Learning}
\label{sec:LfD}

In this section we shortly present the learning and control methods from \cite{suomalainen2018learn6D} that we use for first learning the controller parameters and then performing the actual alignment. For performing the alignment motion we define impedance controllers for both manipulators, separately for translations and rotations, as

\begin{equation}
  \begin{split}
  \pmb{F}_C = K_f(\pmb{x}^*-\pmb{x})+D_f\pmb{v} \\
  \pmb{T}_C = K_\theta(\pmb{\beta}^*-\pmb{\beta})+D_\theta\pmb{\omega}
  \end{split}
  \label{eqt:imp_control}
\end{equation}
where $\pmb{F_C},\pmb{T_C}$ are the commanded forces and torques,  $\pmb{x}^*$ is the desired position, $\pmb{x}$ the current position, $\pmb{\beta}^*$ the desired orientation, $\pmb{\beta}$ the current orientation, $K_f$ and $K_\theta$ stiffness matrices where the axes of compliance are defined and $D_f\pmb{v}$ and $D_\theta\pmb{v}$ linear damping terms.

The trajectory for the master manipulator is computed in a feed-forward manner as
\begin{equation}
  \pmb{x}^*_t= \pmb{x}^*_{t-1}+\pmb{\hat{v}_{d}^*} \nu  \Delta t  \\
\end{equation}
where $\Delta t$ is the sample time of the control loop, $\pmb{\hat{v}_{d}^*}$ the desired direction and $\nu$ the translational velocity. The goal of the method from \cite{suomalainen2018learn6D} is to learn parameters $\pmb{\hat{v}_{d}^*}$, $K_f$ and $K_\theta$ for both manipulators from one or more human demonstrations performed such that the slave manipulator is stiff. 

\begin{figure}[tb]
	\centering
	\includegraphics[width=.70\columnwidth]{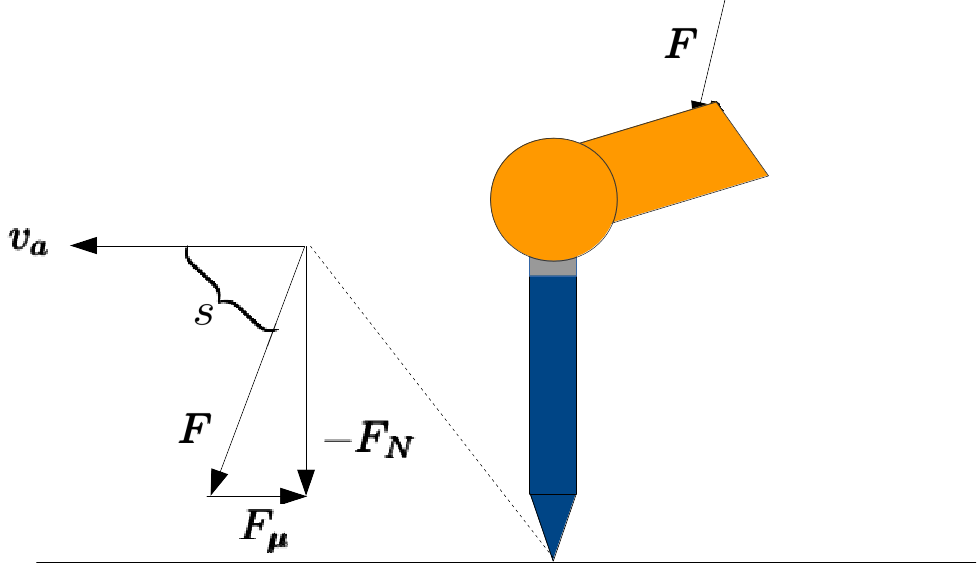}
	\vspace{-0.1cm}
	\caption{\small{Forces acting on the end-effector during a sliding motion. $\pmb{F}$ is the force applied by the controller during a teleoperated demonstration, $\pmb{F}_N$ the normal force, $\pmb{F}_{\mu}$ the friction force, $\pmb{v}_a$ the actual direction of motion and $s$ the sector of desired directions. }}
	\label{fig:sliding_forces}
	\vspace{-0.5cm}
\end{figure}

The intuition for learning the desired direction $\pmb{\hat{v}_{d}^*}$ stems from geometry: to slide the robot's end-effector along a surface, there is always a friction-dependent sector $s$ of directions from which the robot can apply a force to accomplish the sliding. If this sector is calculated at intervals over a whole demonstration, the intersection of all sectors $s_i$ would signify a direction which can lead the end-effector through the whole demonstrated motion either in free space or in contact. We call sector $s$ a set of desired directions and it is visualized for a single time-instant in 2-D in Fig.~\ref{fig:sliding_forces}.

When the manipulator is teleoperated, the sum of environmental forces $\pmb{F}$ from Fig.~\ref{fig:sliding_forces} can be estimated with the desired Cartesian force of the teleoperated arm's controller, which is of similar form as \eqref{eqt:imp_control}. Therefore, as seen from Fig.~\ref{fig:sliding_forces}, the sum of environmental forces for a tool sliding with constant speed can be written as

\begin{equation}
  \pmb{F} = \pmb{F_N} + \pmb{F}_{\mu}
  \label{eqt:measured_force}
\end{equation}
where $\pmb{F}_{\mu}=-\vert\mu\pmb{F_N}\rvert\pmb{\hat{v}_a}$ is the force caused by Coulomb friction with $\mu$ being the friction coefficient and $\pmb{\hat{v}_a}$ a unit vector representing the direction of motion, and $\pmb{F_N}$ the normal force. Similarly, the environmental torque can be estimated from the torque that the controller applies, written as
\begin{equation}
  \pmb{T} = \pmb{r} \times \pmb{F_{\mu}} + \pmb{r} \times \pmb{F_N}
  \label{eqt:measured_torque}
\end{equation}
where $\pmb{T}$ is the controller torque, $\pmb{r}$ the position vector between the measurement point and location of contact. 

In \cite{suomalainen2018learn6D} demonstrations were performed by kinesthetic teaching, i.e. grabbing the robot and leading it through the motions. However, there are cases where kinesthetic teaching is not feasible, for example if the robot is operating in a confined or dangerous workspace. In \cite{suomalainen2018hydraulic} the authors showed that for a 2 degrees of freedom task the method from \cite{suomalainen2018learn6D} can successfully learn compliant motions from teleoperated demonstrations, even though in general providing demonstrations by teleoperation is more difficult than by kinesthetic teaching \cite{fischer2016comparison}. In this paper we show that the method from \cite{suomalainen2018learn6D} can be directly applied to teleoperating a robot manipulator by moving another manipulator, similarly as in \cite{havoutis18AURO}.

The key assumption for detecting the axes of compliance is that if there is motion in other directions besides $\pmb{\hat{v}_{d}^*}$, that motion must be caused by the environment and therefore compliance is required in the direction of that motion. To find the axes, we perform Principal Component Analysis (PCA) on the motion in other directions besides $\pmb{\hat{v}_{d}^*}$ to find out where the most prominent motion besides $\pmb{\hat{v}_{d}^*}$ has occurred. We compute likelihoods of how well each PCA vector fits the data. Based on the likelihoods, we use Bayesian Information Criterion (BIC) \cite{schwarz1978estimating} to decide which of the PCA vectors need to be compliant, thus constructing $K_f$ and $K_\theta$.

Initially, these $K_f$ and $K_\theta$ are learned for the master manipulator. We propose using the same matrices directly for the slave manipulator as well, defined in the slave's \ac{tcs}. We can see from Figs.~\ref{fig:coc}b and \ref{fig:coc}c that both manipulators need to rotate around the same axis according to their respective \ac{tcs}s to succeed in alignment. Additionally, as the workpieces are essentially coupled when touching each other, their translational motion caused by compliance will also occur along the same direction. We argue that the difference in the translational axes of compliance caused by the misalignment of the workpieces is not significant enough to have a major effect on the task. 

\section{EXPERIMENTS AND RESULTS}
\label{sec:experiments}
We performed experiments with a Rethink Robotics Baxter dual-arm manipulator and two Franka Emika Panda manipulators. We used a peg-in-hole setup made of PVC plastic, where the peg has radius 16.5~mm, length 80~mm and a rounded tip, and the clearance with the hole is 0.25~mm. The arrangement is depicted in Fig.~\ref{fig:setup}. The part with hole was attached to the wrist of the right Panda robot (in the view of Fig.~{\ref{fig:setup}) and the peg was attached to the wrist of the left Panda robot, the master robot. Instead of grasping, the peg and hole were rigidly attached to the wrists of the robots to allow better measurements for evaluating the method. 

\begin{figure}[tb]
	\centering
	\includegraphics[width=.9\columnwidth]{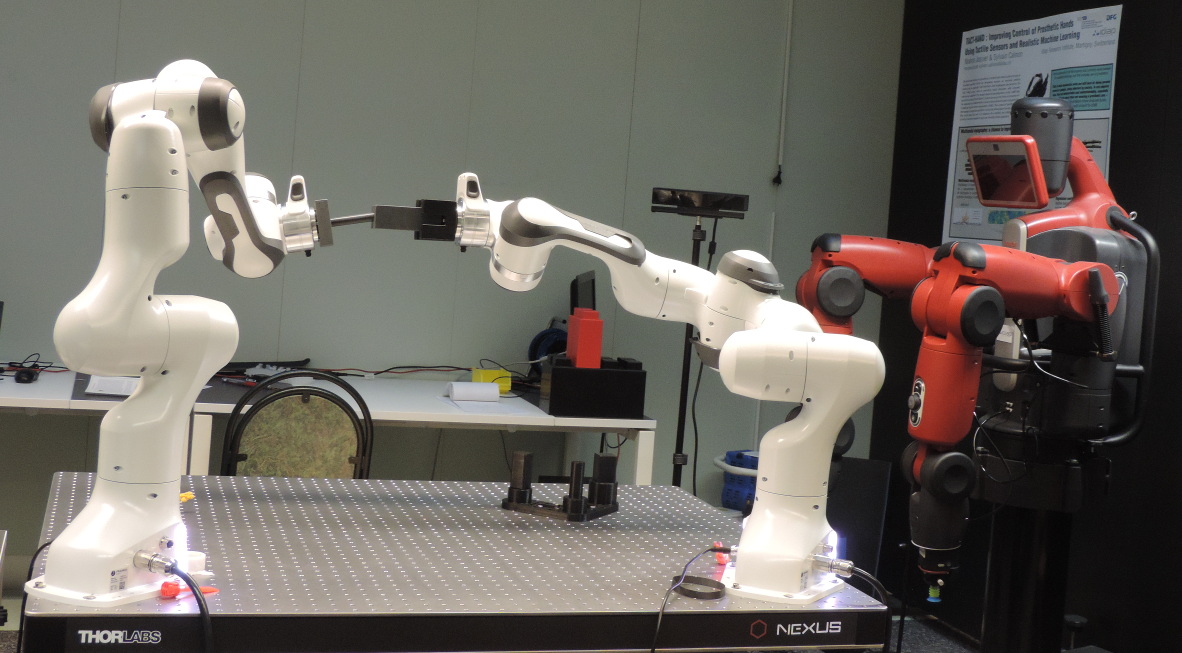}
	\caption{\small{The experimental setup: two Panda robots (peg attached to the left, master, and hole attached to the right, slave) used for performing the task, and the Baxter robot used for teaching by teleoperation.}}
	\label{fig:setup}
\vspace{-1em}
\end{figure}

For the demonstrations, we teleoperated one of the Panda arms with one of the Baxter arms, while the other Panda arm was set stiff. The teacher moved the Baxter arm kinesthetically by using a controller compensating the effect of gravity. The teleoperation was performed with relative motions in \ac{tcs}, i.e. the Panda repeated in its own \ac{tcs} the motions that were performed with the Baxter in the Baxter's \ac{tcs}. Motion from the master Panda was recorded and the environment wrenches were estimated based on the desired controller wrenches. With the method from \cite{suomalainen2018learn6D} we learned the parameters required to perform the peg-in-hole alignment in the case of a stiff slave. The learned desired direction of translation $\pmb{\hat{v}_{d}^*}$ was along z-axis in \ac{tcs}. No desired direction of orientation was required. One axis of compliance in translation detected, along the x axis. One axis of compliance in rotation detected, around the y axis.

The parameters were learned for a \ac{coc} at the wrist only. In \cite{suomalainen2018learn6D} we already analysed the possibilities of learning in different coordinate systems. As these parameters are also what our reasoning in Section~\ref{sec:pih} suggests, for the sake of comparison we use them with all our experiments. However, to get a better idea of which of the parameters are most significant in terms of performance, we chose only some of the parameters for certain tests, but the individual axes were learned from demonstrations. Thus, from now on when we write \textit{compliance in rotation}, we mean compliance around y axis as learned, and similarly for translation. By varying the number of compliant axes, we studied the following five compliance configurations:
\begin{enumerate}[label=\emph{\Alph*}]
\item Compliance in translation and rotation in master's wrist only, similar to teaching.
\item Compliance in translation and rotation in master's tooltip only.
\item Compliance in translation and rotation in slave's tooltip only.
\item Compliance in rotation only, both master and slave tooltip.
\item Compliance in rotation and translation in both master and slave tooltip.
\end{enumerate}

\subsection{Convergence region}
With the term \textit{convergence region} we mean the region from which reproduction with the same set of parameters is possible. More specifically, in these experiments we mean the maximum orientation error in the rotation around y axis in \ac{tcs} from which the reproduction is successful. Even though the resulting values are specific for our setup, we can use this measure to compare configurations \textit{A-E}. Table~\ref{tab:angles} shows the largest angles for which the alignment was successful with each set of parameters. During the experiment the orientation error was increased with 1 degree resolution from 4 degrees until the reproduction failed. Each experimental condition was repeated 5 times. In all cases all 5 repetitions were either successful or unsuccessful, which is why measurement uncertainty is not presented in Table \ref{tab:angles}. 

\begin{table}[tbp]
  \centering
  \begin{tabular}{@{}cccccc @{}}
  	\toprule 
  	{\bf Configuration} & {\emph{A}} & \emph{B} & \emph{C} & \emph{D} & \emph{E} \\ \midrule
        {\bf Max.~orientation error (deg)} & {4} & 12 & - & 8 & 12\\  \bottomrule
  \end{tabular}
  \caption{\small{A table summing up the maximum orientation errors with which the alignment was successful. In case \emph{C} the reproduction failed even with 4 degrees error.}}
  \label{tab:angles}
  \vspace{-0.5cm}
\end{table}

The first choice was the location of \ac{coc}, i.e.~comparison between experiments \emph{A} and \emph{B}. It was observed that compliance in tooltip has a more significant effect on the convergence region, as expected after the analysis of Section~\ref{sec:coc}. After this result, we performed the rest of the experiments with \ac{coc} at tooltip.

In condition \emph{C}, only the slave was set compliant, with the \ac{coc} at tooltip. However, with this setting even 4 degrees error prevented the motion. This is contradictory to our analysis in Section~\ref{sec:coc}. A probable explanation is the internal friction at the robot joints. Since there are already forces applied at the joints of the master, it requires only a small amount of additional force to cumulatively overcome the static friction. However, in the slave manipulator, overcoming the static friction would require higher forces, and thus in practice the compliance in the master facilitates the alignment more than compliance in the slave. 

Finally we set both manipulators compliant at tooltip, with the parameters learned earlier used for the respective \ac{tcs} of each manipulators. First we set both master and slave manipulators compliant in rotations only (condition \emph{D}), and then both manipulators to compliant in both translation and rotation (condition \emph{E}). It can be observed that translational compliance is important in this setting, and setting both robots to compliant in both rotation and translation increased tolerance to errors in the starting orientation. However, having both manipulators compliant did not achieve larger region of convergence than having only master manipulator compliant (conditions \emph{B} and \emph{E)}.

We conclude that to maximize robustness against positional errors in an alignment task, at least the master arm should be compliant in both translation and rotation at tooltip. The translational compliance seems to play a key role, since rotational compliances only in both master and slave did not ease the alignment as much, even though when the coordinate system is set to tooltip there is only rotational error, not translational. In the next section we investigate how the motions are affected when also the slave is compliant.

\subsection{Motion analysis}
We recorded the joint positions, forces and duration of the motion during reproduction.
Each of these criteria can present certain advantages.
If the workspace of the robot is limited, it would be advantageous that the robot can perform the motion in a small space. Minimizing the joint motions can also help keep the robot operational for longer time. Performing the motions faster would also be an advantage. Finally, keeping the forces lower decreases the probability of breaking the workpieces being aligned. 

We observed, as was already shown by Zhang et al.~\cite{zhang2017peg}, that having both arms compliant reduces the amount of forces applied to the end-effector. As they did a thorough numerical validation of this, we will not present these results from our experiments for brevity but concentrate on the metrics which they did not analyse. 

We compared the metrics of conditions \emph{B}, \emph{D} and \emph{E} at 8 degrees error since this was the maximum orientation error where the motion was successful in all these three conditions. From 5 repetitions for each condition, we computed the means of the following metrics, summed over all joints: total covered joint distance (rad), total covered joint distance weighed by link lengths, joint covered distance maximum (the most motion by a single joint) and movement duration (seconds). The results are shown in Table \ref{tab:8deg_metrics}. When applicable, the distances of the two arms are presented separately.

It can be seen that having compliance in both manipulators (conditions D and E) produces superior results across all measures. In addition, condition E outperformed condition D, showing that translational compliance plays a significant role in an alignment task. Furthermore, we empirically observed that having both manipulators compliant greatly eased the final push when the workpieces were nearly aligned. 

\begin{table}[tbp]
  \centering
  \begin{tabular}{@{}ccccccc @{}}
  	{} & {} & \multicolumn{5}{c}{\bf Configuration}\\
    {} & {} & \emph{B} & \multicolumn{2}{c}{\emph{D}} & \multicolumn{2}{c}{\emph{E}} \\ \cmidrule{2-7}
    {} & {} & {} & {M} & {S} & {M} & {S}  \\ \cmidrule{2-7}
    {} & \emph{Joint dist} & {1.45} & {0.64}& {0.35} & {0.41} & {0.18} \\ \cmidrule{2-7}
    {\bf Metric} & \emph{W. joint dist} & {1.18} & {0.53}& {0.23} & {0.36} & {0.15}\\ \cmidrule{2-7}
    {} & \emph{Max.~joint dist} & {1.1} & {0.47}& {0.19} & {0.32} & {0.13}\\ \cmidrule{2-7}
    {} & \emph{Time} & {26.5} & \multicolumn{2}{c}{13.0} & \multicolumn{2}{c}{5.00} \\  \cmidrule{2-7}
  \end{tabular}
  \caption{\small{Various metrics evaluated for a successful alignment task with 8 degrees rotational error. M stands for master manipulator and S for slave, W.~means weighed with the link lengths and Max.~the maximum norm} }
  \label{tab:8deg_metrics}
  \vspace{-0.5cm}
  
\end{table}

\subsection{Discussion}
\label{sec:discussion}
In the results of Table~\ref{tab:angles} the failure type that confined the convergence region was always jamming (Fig.~\ref{fig:failmodes}a). In the initial trials we also observed failures according to Fig.~\ref{fig:failmodes}c when the orientation error was considerably higher than the maximum orientation error in Table~\ref{tab:angles}. Thus we reason that failure type presented in Fig.~\ref{fig:failmodes}c occurs either with high orientation error or low friction.

It should also be noted that the set of parameters learned from the human demonstration were the ones that produced best results in the experiments when applied to both manipulators. This experiment showed that the parameters can be learned even when the demonstrations are performed with a difficult teleoperation method. Finally, we also validated that with the same parameters the task can also be completed when the error angle is in the other direction (i.e.~the master is sliding "up").


\section{CONCLUSIONS AND FUTURE WORK}
\label{CONCLUSION}
We presented an analysis on how to choose the center and axes of compliance for a dual-arm alignment task where the master moves along a linear trajectory and both arms can be set to compliant. In addition, we showed how to easily learn these parameters from a human demonstration performed with one teleoperated manipulator only. A slightly unexpected result was that adding compliance to the slave manipulator in addition to the master did not increase the convergence region. We suspect the reason to be similar that prevented alignment with compliance in slave manipulator only: due to the static joint frictions, the master manipulator must move first and the slave compliance will only have an effect later in the motion. We showed through experiments that slave compliance in addition to the master had other significant benefits, namely shorter joint motions and time. 

In this paper we concentrated more on the rotational compliance, the main reason being that the assumed error was in rotation. However, the results showed that translational compliance is a major factor in easing the alignment, even though the analysis from Section \ref{sec:pih} suggests that rotational compliance suffices. The exact mechanics for the role of translational compliance would require further mechanical analysis of the system.

An interesting concept would be to have the \ac{coc} of the slave match that exactly of the master---in other words, make the \ac{coc} change dynamically during the motion according to the pose of the master. Our experimental setup did not allow a dynamically changing \ac{coc}, but the experiment that produced the best results, having the \ac{coc} of both robots at the tooltip is an approximation of this. 



\bibliographystyle{ieeetr}
\bibliography{biblio}

\end{document}

%% file: acronyms.tex
\newacro{tcp}[TCP]{Tool Center Point}
\newacro{lfd}[LfD]{Learning from Demonstration}
\newacro{coc}[CoC]{Center of Compliance}
\newacro{ft}[F/T]{Force/Torque}
\newacro{tcp}[TCP]{Tool Center Point}
\newacro{tcs}[TCS]{Tool Coordinate System}